\documentclass[letterpaper, 10 pt, conference]{ieeeconf}  

\IEEEoverridecommandlockouts                              

\usepackage{amsmath} 
\usepackage{amssymb}  
\usepackage{dsfont}
\usepackage{graphicx}

\usepackage{psfrag,graphicx,epsfig}
\usepackage{epstopdf}
\usepackage{xspace}
\usepackage{subfig}
\usepackage{float}
\usepackage{placeins}
\usepackage{multirow}
\usepackage{pgf,tikz}
\usepackage{nowidow}
\usepackage{lineno}
\usepackage{xcolor}

\usepackage{siunitx}
\usepackage{color}
\usepackage{flushend}
\usepackage{lineno}
\usepackage{algorithmic}
\allowdisplaybreaks

\title{\LARGE \bf
Imitation and Supervised Learning of\\ Compliance for Robotic Assembly 
}

\author{Devesh K. Jha, Diego Romeres, William Yerazunis and Daniel Nikovski$^{\dagger}$%
\thanks{$^{\dagger}$All authors are with Mitsubishi Electric Research Laboratories (MERL), Cambridge, MA, USA 02139 {\tt\small \{jha,romeres,yerazunis,nikovski\}@merl.com}}}%

\begin{document}
\maketitle
\thispagestyle{empty}

\begin{abstract}
We present the design of a learning-based compliance controller for assembly operations for industrial robots. We propose a solution within the general setting of learning from demonstration (LfD), where a nominal trajectory is provided through demonstration by an expert teacher. This can be used to learn a suitable representation of the skill that can be generalized to novel positions of one of the parts involved in the assembly, for example the hole in a peg-in-hole (PiH) insertion task. Under the expectation that this novel position might not be entirely accurately estimated by a vision or other sensing system, the robot will need to further modify the generated trajectory in response to force readings measured by means of a force-torque (F/T) sensor mounted at the wrist of the robot or another suitable location. Under the assumption of constant velocity of traversing the reference trajectory during assembly, we propose a novel accommodation force controller that allows the robot to safely explore different contact configurations. The data collected using this controller is used to train a Gaussian process model to predict the misalignment in the position of the peg with respect to the target hole. We show that the proposed learning-based approach can correct various contact configurations caused by misalignment between the assembled parts in a PiH task, achieving high success rate during insertion. We show results using an industrial manipulator arm, and demonstrate that the proposed method can perform adaptive insertion using force feedback from the trained machine learning models.
\end{abstract}

\section{Introduction}
\label{sec:Introduction}

\begin{figure}
    \centering
    \includegraphics[width=0.38\textwidth]{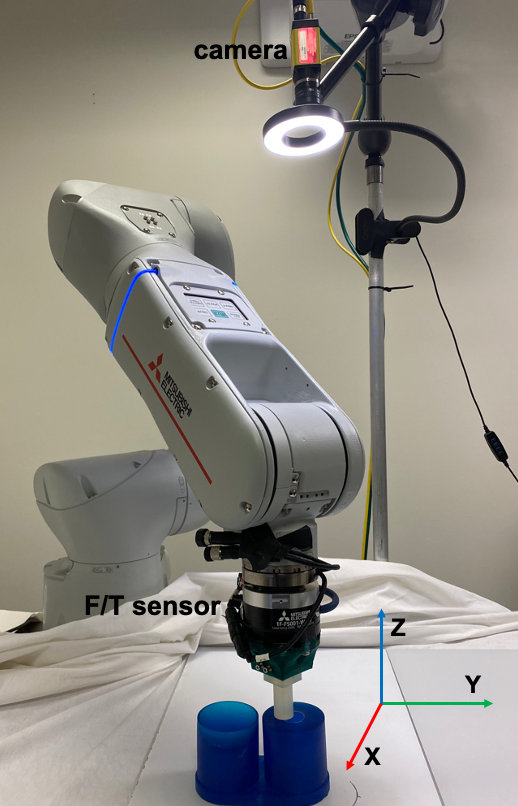} %
    \caption{Experimental setup with a Mitsubishi Electric Factory Automation (MELFA) RV-$5$AS-D Assista $6$  DoF manipulator arm in a possible contact configuration with the hole environment. The diameter of the peg is approximately $20$ mm and the tolerance is approximately $1.5$ mm. The figure also shows the industrial vision system that was used in the experiments for detecting the hole.}
    \label{fig:exp_setup}
\end{figure}
Mating parts under tight tolerances (see Figure~\ref{fig:exp_setup}) is one of the most important operations in robotic assembly. Over the last several decades, robots have become very precise in performing pick and place operations. However, complications arise when the positions of the parts involved in assembly vary between repetitions of the operation. This can happen in cases where the parts are deposited onto a surface by a feeder, and each time end up in a different position; also, in a case where the parts arrive on a moving conveyor belt. In such cases, industrial vision cameras can be used to determine the parts’ position. However, the determined position of the parts by the industrial vision cameras is usually fairly inaccurate, often by several millimeters. This is significantly more than the required tolerances during assembly. For this reason, standard compliance (e.g., stiffness) controllers usually fail to perform the operation successfully. Besides, even if the robotic device knows the exact position of the parts, its path (via points) still needs to be modified in order to accommodate the variation in the position of the parts. In practice, such a modification is performed by means of dedicated software that takes as input the changed position of the parts and outputs a new path for the robot. However, developing such dedicated software is typically very difficult and laborious, and is currently one of the main contributors to the high cost of deploying and re-tooling robotic devices for new assembly operations.

Therefore, there is a strong industrial need for methods for creating robot controllers that can accommodate varying positions of assembly parts with minimal or no programming. That is, the robot controller should be able to generalize over the positions of the parts. Arguably, generalization is one of the hallmarks of intelligence, be it human or artificial, and endowing robots with the ability to generalize over the starting conditions of assembly tasks would amount to a qualitatively different type of intelligent assembly functionality. Such a functionality would match the ability of human workers to learn a new assembly skill and apply it to many new situations.  

In this paper, we present a two-fold approach for assembly so that the agent can generalize to a different initial condition and accommodate uncertainty in the final goal position. A demonstration is used to learn the approach movement of the robot from any initial position to a novel target position. Then, a supervised learning method is used to learn a force-feedback policy to accommodate and correct for the uncertainty in the estimated target position. The purpose of the force-feedback policy is to interpret the contact information and correct the possible misalignment. A commonly used prototypical robot assembly operation is the insertion of a peg into a hole (see Figure~\ref{fig:exp_setup}), and we will use this task as an example when describing and evaluating the algorithm. We present a method that allows a robot equipped with a camera and a force/torque sensor mounted on its wrist to learn this operation, with minimal or no programming, even when the estimates of the positions of the parts (returned by the camera) are not very precise. We assume that a human instructor can provide at least one demonstration of a successful operation, using the robot itself.  

\begin{figure}
    \centering
    \includegraphics[width=0.45\textwidth]{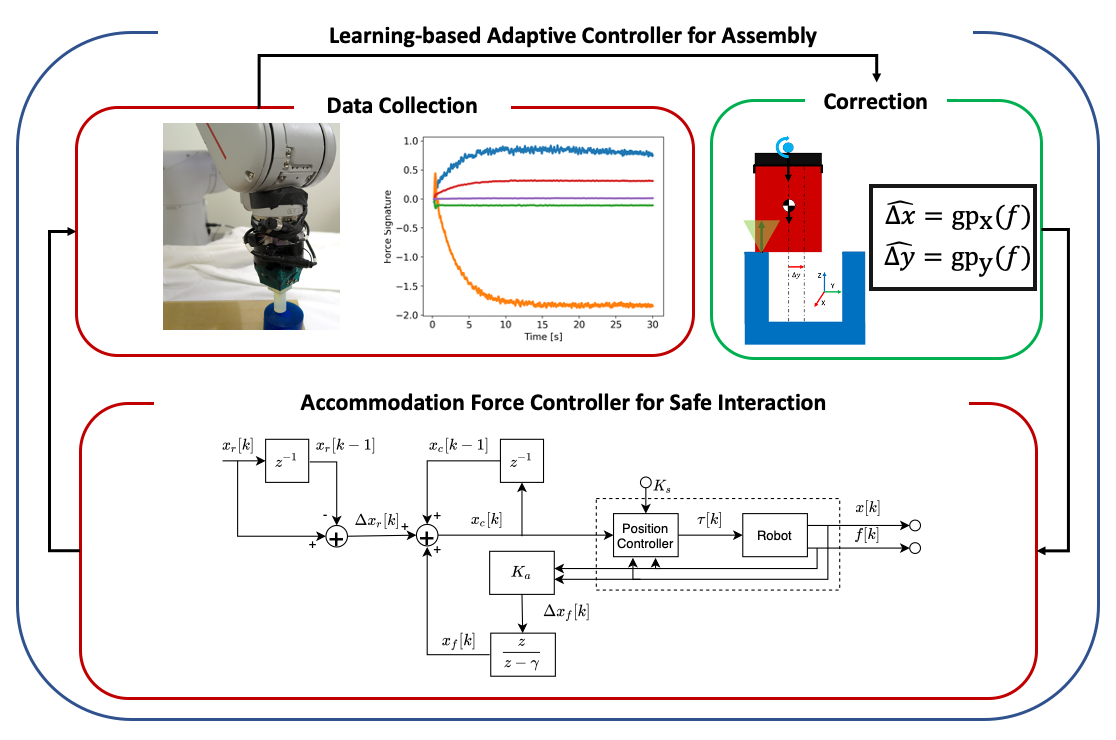} %
    \caption{The proposed method for design of a force controller-based learning method for part assembly. The proposed method uses an accommodation controller as a low-level controller for safe interaction during data collection. The data collected during exploration is used to train regression models to correct for the misalignment between the peg and the hole. This machine learning model-based controller is used at the higher level to correct the misalignment, so as to perform insertion.}
    \label{fig:proposed_method}
\end{figure}

\textbf{Contributions:} Figure~\ref{fig:proposed_method} provides an abstract idea of the proposed method for peg-in-hole assembly. To enable safe sustained contacts with the environment during insertion attempts, we present the design of an accommodation controller. We use our novel accommodation controller as the low-level controller during exploration as well as operation. The proposed method uses this accommodation controller to maintain safe contact during the data collection process. A machine learning model then learns a mapping from the 
(quasi-)static contact wrench to the desired correction to perform insertion. In particular, the proposed paper has the following contributions:
\begin{itemize}
    \item We present a design of a learning-based adaptive controller to perform insertion with incorrect knowledge of the goal location, which can achieve near-perfect insertion performance. This is demonstrated on a physical robotic system with industrial-grade vision systems.
    \item We present the design and analysis of a stable accommodation controller that allows maintaining contacts with the external environment while ensuring contact forces to always stay within desired bounds when following a dynamic reference trajectory.
    
\end{itemize}

\section{Related Work}
\label{sec:RelatedWork}
Research on robotic assembly has a long history, and a lot of this research has used PiH insertion as a prototypical task. The main reason this task is difficult to automate is that when the peg and hole are in contact, even a small misalignment between their correct positions would cause a controller that tracks a desired trajectory defined in terms of robot positions or applied torques at the joints to result in very large contact forces that might break the robot and/or the parts. What is needed is a method that adjusts the motion of the parts in response to contact forces. Such techniques are also known as {\em adaptive assembly strategies} (AAS) (\cite{Nevins1977ResearchAutomation,Mason1981ComplianceManipulators,Inoue1974ForceTasks,Gullapalli1994AcquiringLearning,Abu-Dakka2015AdaptationProfiles,Levine2016End-to-endPolicies,dong2021icra,lee2020making,Kulkarni2021LearningLearning, DBLP:journals/corr/abs-2007-11646}). 

It is generally desirable that a controller be able to make an active interpretation of force/torque (F/T) signals to generate corrective actions.  Several ideas have been proposed in the literature to design controllers to compensate for larger misalignment during insertion. The idea in many of these controllers is to use a mapping from the contact wrench measured at the wrist of the robot to a correction in the motion for the robot. For PiH problems, in rare instances, it is possible to compute this mapping analytically --- for example, when the axes of a round peg and hole are aligned perfectly, the point around which moments measured by the F/T sensor lies on axis of the peg, and at least some overlap between the peg and hole exists, it is possible to calculate the direction to the center of the hole using a closed-form expression (\cite{Gottschlich1989AMating}). However, this solution requires careful calibration of the sensing system, as changing the location of the peg with respect to the end tool of the robot and its F/T sensor will affect the sensed force signature in a given contact configuration and render the analytical formulae incorrect. 

Another approach for obtaining the mapping is to use machine learning algorithms, such as supervised learning or reinforcement learning. An early method for {\em programmed compliance} proposed in \cite{Peshkin1990ProgrammedAssembly} used linear mappings in the form of linear admittance or accommodation matrices, much like earlier compliance (e.g. stiffness) controllers. Later, it was demonstrated that for insertion tasks, the set of specified appropriate examples of corrective action cannot be represented by a linear mapping, in general, and the use of nonlinear regression models such as neural networks was proposed (\cite{Asada1990TeachingCompliance,Asada1993RepresentationNets}). This further significantly expanded the representation power of the learned controller, but the question remained how to come up with this set of examples of correct response to contact forces. To this end, a reinforcement learning (RL) solution based on a process of trial and error was proposed in~\cite{Gullapalli1994AcquiringLearning}, leading to a very impressive solution of a PiH problem with tight tolerances that were orders of magnitude tighter than the accuracy (repeatability) of the used robot. However, this method still assumed accurate knowledge of the goal state, so it is not directly applicable to the problem of insertion with uncertain goal state.

Following the seminal work in~\cite{Gullapalli1994AcquiringLearning}, there have been numerous works using machine learning models for designing adaptive controllers. End-to-end visuomotor training of robots by means of deep RL was demonstrated in \cite{Levine2016End-to-endPolicies,lee2020making}. More recently, there has been some work on using tactile sensors for designing feedback policies to perform insertion~\cite{Dong2019Tactile-BasedBox-Packing, dong2021icra}. The use of RL with recurrent neural networks for insertion problems was demonstrated recently in \cite{Kulkarni2021LearningLearning}. However, although being a very powerful general method for learning control policies, RL is also known for its unfavorable sample complexity, making it a less attractive choice for learning on actual physical systems. Compared with all these works, we present an approach that uses a combination of LfD and a supervised learning-based method for local adjustment to the LfD trajectory that is very sample-efficient in its interaction with the physical equipment. Furthermore, we also present the use of a special-purpose accommodation controller to safely interact with the environment during exploration for data collection and actual insertion operations. 
\section{Problem Statement}
\label{sec:proposed_problem_statement}

Peg-in-hole insertion is an entire class of problems whose difficulty can vary widely depending on the type of misalignment present (positional only or positional plus directional), its magnitude (whether the peg is already partially in the hole, or if not, whether the peg and hole at least overlap some), the tolerance of the insertion operation, whether the parts are chamfered, what kind of sensors are used (e.g., visual, force, tactile), what their relative accuracy is, etc. (\cite{Whitney1987HistoricalControl}). An additional consideration is how the robot is controlled (full joint torque control or position control only). 

In this paper, we are considering a version of the problem that corresponds to fairly typical circumstances often encountered in practice. We are designing controllers to handle inaccuracies in the goal position that are much larger than the tolerance of insertion, but still smaller than the size of the parts being inserted. This situation arises, for example, when we use a camera to estimate the position of the hole. The positional estimation error of modern cameras for observed objects in the world frame tends to be on the order of several millimeters, which is much larger than typical insertion tolerances in assembly (often sub-millimeter), but still less than the radius of the parts being inserted. This assumption entails that if the robot executes in open loop a motion to where it believes the goal state is, even if the insertion does not succeed and the peg collides with the rim of the hole, the resulting contact configuration will have the peg and hole overlap to at least some degree, allowing the measured force signature to be used to correct the estimate of the goal position. (If this condition is not true, and there is no overlap, a single F/T reading cannot disambiguate the contact configuration, and the controller must resort to integrating multiple readings, or perform search; this is a much more difficult version of the problem that we are not exploring in this paper, and is usually addressed by very different kinds of algorithms in the research community.) In addition, in this paper, we are assuming that there is no (or very minimal) directional misalignment between the peg and the hole, which corresponds to the case when the hole moves on a plane (for example, a work surface in a factory), changing only its position, but not the angle of its axis with respect to the plane.

In terms of contact sensors, we are considering a force/torque sensor mounted at the wrist of the robot. Furthermore, we are using industrial manipulators that are controlled only in (possibly compliant) position-control mode, that is, we do not assume the availability of direct joint-torque-control mode. This version of the problem matches almost exactly the one used in recent work reported in \cite{Kulkarni2021LearningLearning}, as well as earlier work on robotics assembly \cite{Gottschlich1989AMating}.


\section{Proposed Method}
The method proposed in this paper uses two different machine learning methods, as described below. One of these methods is to learn a nominal trajectory for insertion in the form of a dynamical movement primitive (DMP, \cite{Ijspeert2013DynamicalBehaviors}), and adjust it appropriately for novel start and goal states. The second method is based on a form of active supervised learning that collects training examples relating various displacements of the peg from its correct position for insertion and the F/T observations measured as a result, by means of active experimentation with the peg and hole. 



Learning starts by collecting one or more demonstrations of the operation by an expert. The result of the demonstration(s) is one or more trajectories in Cartesian space of the end tool of the robot of the form $\mathbf{x}(t)$, $t \in [0,T]$, $\mathbf{x}(t) \in \mathbb{R}^3$. Pose of the parts can be estimated by means of one or more cameras. Pose estimation by means of computer vision is an active area of research (\cite{Hodan2018APose}), and various methods provide different trade-offs between cost, speed, and accuracy. We assume that there exists a pose estimation module which is properly calibrated with respect to the start and end of the demonstration trajectory $\mathbf{x}(t)$, that is, $\mathbf{x}_0=\mathbf{x}(0)$ and $g=\mathbf{x}(T)$ are indeed the estimates returned by the camera.



After one or more trajectories $\mathbf{x}(t)$ have been recorded, a dynamic movement primitive (DMP) learning algorithm is used to learn a separate DMP for each of the components of $\mathbf{x}(t)$ (\cite{Ijspeert2013DynamicalBehaviors}). The resulting set of DMPs can generate a new desired trajectory $\mathbf{x}_r(t)$, given a new goal position $g_d$, by integrating the DMP forward in time. This kind of generalization over the goal state is an essential function of DMPs. If the new desired goal state $g_d$ was indeed the correct one, then following the new desired trajectory $\mathbf{x}_r(t)$ would presumably reach the new goal state. However, the estimate of the desired new goal state $g_d$ is usually incorrect, because this estimate depends on the new position of the hole, as described below. Any error in the estimate of the position of the hole translates into an error in $g_d$. For this reason, the control law executed by the robot needs to provide compliance. We use a control law of the form $\mathbf{x}_c(t)=\mathbf{x}_r(t)+H(f)$, where $f$ is the vector of sensed forces and torques by the FT sensor, and $H$ is a non-linear mapping that produces corrections to the position of the robot.  Here, $H$ is a general non-linear mapping, which is learnt using a self-supervised learning method~\cite{Groome1972ForceThesis,Whitney1977ForceMotions,Peshkin1990ProgrammedAssembly,Asada1990TeachingCompliance}. 


The purpose of the next stage of the algorithm is to learn the mapping $H$ through self-experimentation. To this end, while the hole is still in its original position, variations $d(t)$ are introduced to one of the original trajectories $x(t)$ demonstrated by a human operator that successfully completed the operation. To learn this mapping, we collect force signature data of the form $(f_i, d_i)$, where $f_i$ is the force wrench experienced by introducing a variation $d_i$.  However, based on the understanding that the mapping from displacements to forces is typically many-to-one (multiple displacements would sometimes result in the same force), following the analysis in \cite{Newman2001InterpretationAssembly}, the inverse mapping would be one-to-many, that is, not a function that can be learned by means of machine learning. However, it can be realized that the exact magnitude of the displacement does not need to be recovered for successful corrective action, and furthermore, multiple displacements can generate the same forces only if the sign of these displacements is the same, as long as the magnitude of the displacement does not exceed the radius $R$ of the object being inserted. Based on this realization, a supervised machine learning algorithm can be used to learn the mapping  $sign(d_i)=H_0 (f_i)$, for all examples $i=1,N$ such that $|d_i |\leq R$. Such a learned machine learning model is then used to a design a corrective controller to correct the misalignment between the mating parts.


A practical consideration about the exploration method is how to do it safely. If the robot is commanded to follow the modified trajectory in standard position control mode, the resulting collision will either damage the robot or the parts, or result in a fault due to excessive force. This means that the robot needs to be controlled in a compliant control mode. One accommodation controller design for safe exploration and exploitation of the learned control policy based on these principles and suitable for execution on industrial robot arms is descried in the next section. 

\begin{figure*}
    \centering
    \includegraphics[width=0.75\textwidth]{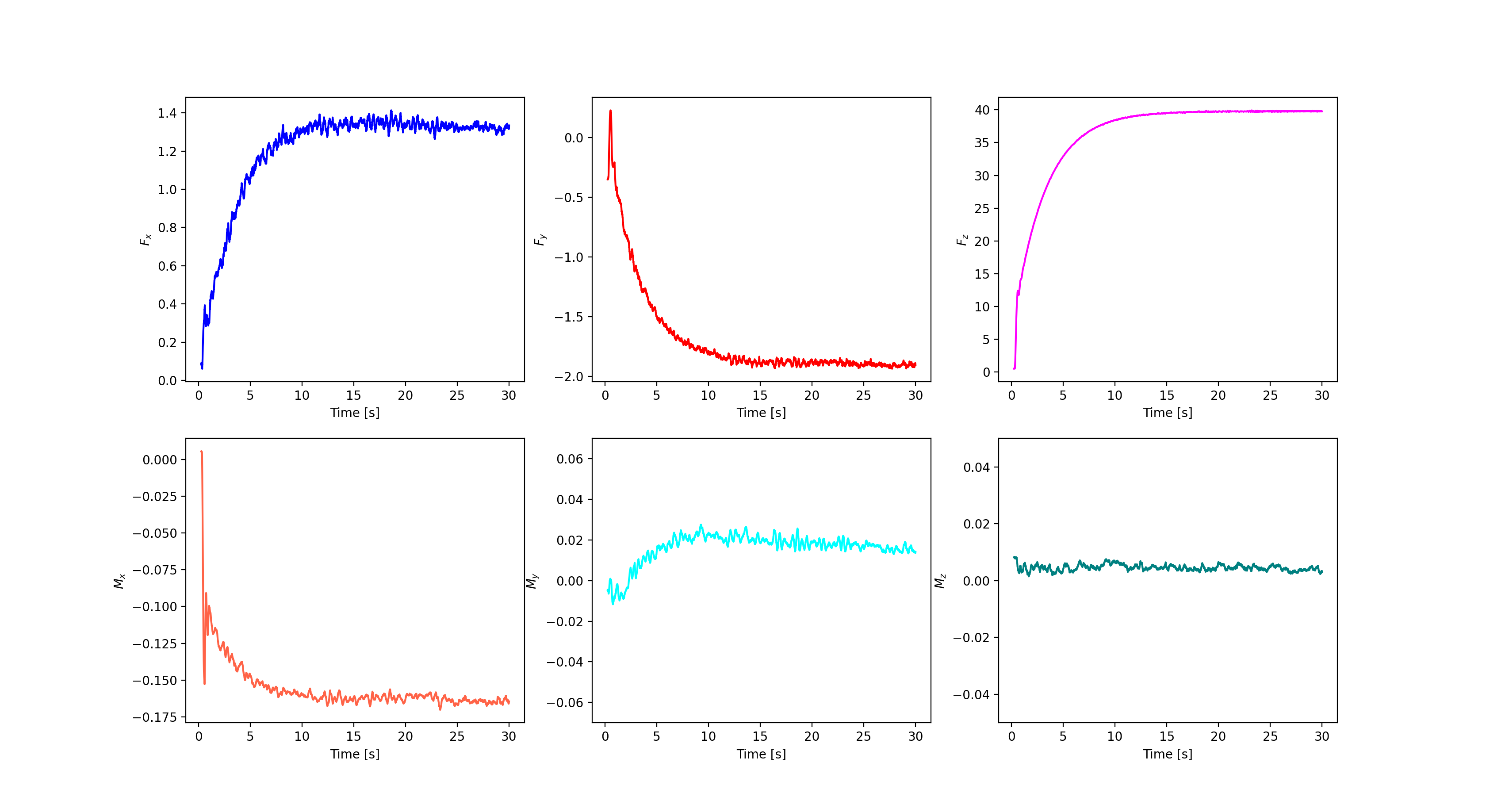} %
    \caption{This plot shows the convergence of contact wrench obtained using the proposed GAC during a contact configuration during an insertion attempt. The reference trajectory advances linearly in the insertion direction at constant velocity even after actual movement of the robot stops due to collision of the peg with the rim of the hole. For the same reference trajectory, an admittance controller would have produced ever increasing forces and moments.}
    \label{fig:accommodation_controller_plot}
\end{figure*}

\section{Force Controller Design for Safe Interaction}\label{sec:accommodation_controller_design}

\begin{figure}
    \centering
    \includegraphics[width=0.50\textwidth]{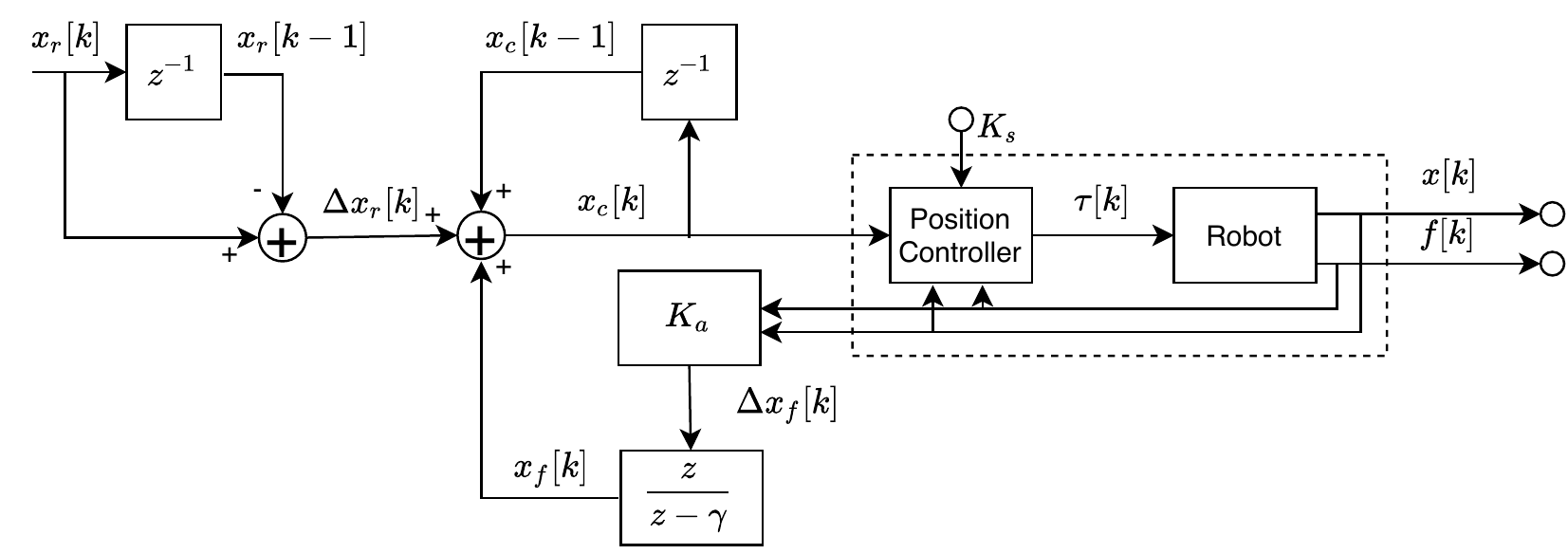} %
    \caption{Block Diagram of the generalized accommodation controller that is designed for the data collection process. $K_a$: Accommodation Matrix, $K_s$: Stiffness Gains of the low-level stiffness controller.}
    \label{fig:gen_acc_block_diagram}
\end{figure}

While there has been significant work in the literature to perform active control using the contact forces during insertion-type assembly operations, design of controllers to ensure safe, sustained contacts has received less attention. In this section, we present the design and analysis of a force controller that ensures safe interaction by maintaining contact forces within safe bounds. In particular, we try to answer and explain the following in this section:
\begin{enumerate}
    \item Design of a Generalized Accommodation Controller (GAC) for safe exploration during assembly attempts.
    \item Evaluate the convergence characteristics of the proposed controller.
\end{enumerate}
It is important to note that the design of GAC is critical to the design of the ACC, as this allows us to collect data in quasi-steady state by making the contact forces converge to a value that is representative of the contact configuration, and not of the applied effort by the robot, thus allowing interpretation of the phenomena in quasi-static states (instead of in dynamic states). It is also desirable to have such a controller to make sure that the interaction force between the robot and its environment stays within some desirable safe bounds. We first explain the design of this controller using the stock stiffness controller provided in general industrial robots. 

\subsection{Generalized Accommodation Controller Design} The idea of the proposed accommodation controller is presented in Figure~\ref{fig:gen_acc_block_diagram}. As could be seen in the block diagram in Figure~\ref{fig:gen_acc_block_diagram}, the accommodation controller manipulates the reference trajectory using force feedback. In particular, the accommodation controller uses the following feedback law to manipulate the reference trajectory of the robot. Let us denote the discrete-time reference trajectory by $x_r[k]$, the trajectory commanded to the low-level position controller by $x_c[k]$, the experienced forces by $f[k]$, the measured position as $x[k]$, at any instant $t_k$. Let us denote the stiffness constant of the compliant position controller by $K_s$ and the accommodation matrix for the force feedback as $K_a$. For simplicity, we consider a diagonal matrix $K_a$. The commanded trajectory sent to the robot is then computed using the following update rule:
\begin{equation}
      x_c[k]=x_c[k-1]+\Delta x_r[k]+\sum_{i=0}^k\gamma^{k-i}K_a f[k]
\end{equation}
where $\gamma \in (0,1)$ is a discounting parameter for computing the integral error, and $\Delta x_r[k]=x_r[k]-x_r[k-1]$ are desired position increments computed from the reference trajectory (normally itself computed by the DMP module based on its estimate of the goal position). Note that the force experienced during interaction is governed by the stiffness constant of the robot controller and is given by $f[k]=K_s(x[k]-x_c[k])$. We assume that such a controller is available in the control software of the robot; if not, it can be implemented by interpreting the contact forces accordingly (\cite{Lynch2017ModernRobotics}).

\subsection{Convergence Behavior of the Accommodation Controller}
In this section, we briefly describe the convergence behavior of the accommodation controller that was described in the previous section. It has the useful property that even when the reference trajectory advances indefinitely, the contact force converges to a constant, as long as the reference velocity is constant. Because the reference velocity (that is, the velocity at which the desired reference trajectory $x_r[k]$ is traversed) is entirely under the control of the designer of the controller, this effectively means that the contact force is under their control, too, and can be set to a desired safe value. This is in contrast to the behavior of a standard compliance controller, such as a stiffness controller, where contact forces would increase indefinitely if the reference trajectory keeps advancing for any reason.

Convergence of the GAC can be understood by considering that by using suitable values for the accommodation matrix $K_a$ and the discounting factor $\gamma$, the contact forces can be regulated so that both they and the commanded trajectory reach a constant value, when the position change in the reference trajectory is kept constant, that is, $\Delta x_r[k]=v$ for some constant velocity $v$ along the intended direction of insertion. To see this, the commanded position will stop changing ($x_c[k]=x_c[k-1])$ when both update terms to it cancel each other, that is, $v=\Delta x_r[k] =-\sum\limits_{i=0}^{k}\gamma^{k-i}K_a f[k]$. Under this condition, the commanded trajectory converges to $\tilde{x}_c=x_c[k]$. Then, if the movement of the robot has also stopped at some equilibrium position $\tilde{x}$, under the operation of the stiffness position controller of the robot with a constant commanded position $\tilde{x}_c$, the steady-state force experienced by the robot is given by $\tilde{f}=K_s(\tilde{x}-\tilde{x}_c)$. Then, at steady state, $v=K_a/(1-\gamma)\tilde{f}$, or $\tilde{f}=v (1-\gamma)/K_a$. This demonstrates that the contact force at equilibrium can be reduced by choosing a lower speed of traversal of the reference trajectory, or a higher discounting factor $\gamma$, or a lower value of the accommodation constant $K_a$. The convergence behavior of the forces is also shown in Figure~\ref{fig:accommodation_controller_plot}. As could be seen in the plots, the force and moments experienced by the robot converge to a steady-state value. The speed of convergence of the system is governed by the discounting parameter $\gamma$. We observed that the convergence is faster with higher values of $\gamma$. In practice, any parameter $\gamma \in [0.3,0.65]$ could achieve satisfactory convergence. Values of $\gamma$ higher than $0.65$ might sometimes lead to oscillations in the system.

In the rest of the paper, we use this controller as the trajectory-following controller which is used when following a reference trajectory obtained from a higher level controller working towards achieving successful insertion for assembly. This is described in detail in next section.
\section{Experimental Results}\label{sec:results}
\begin{figure*}
    \centering
    \includegraphics[width=0.8\textwidth]{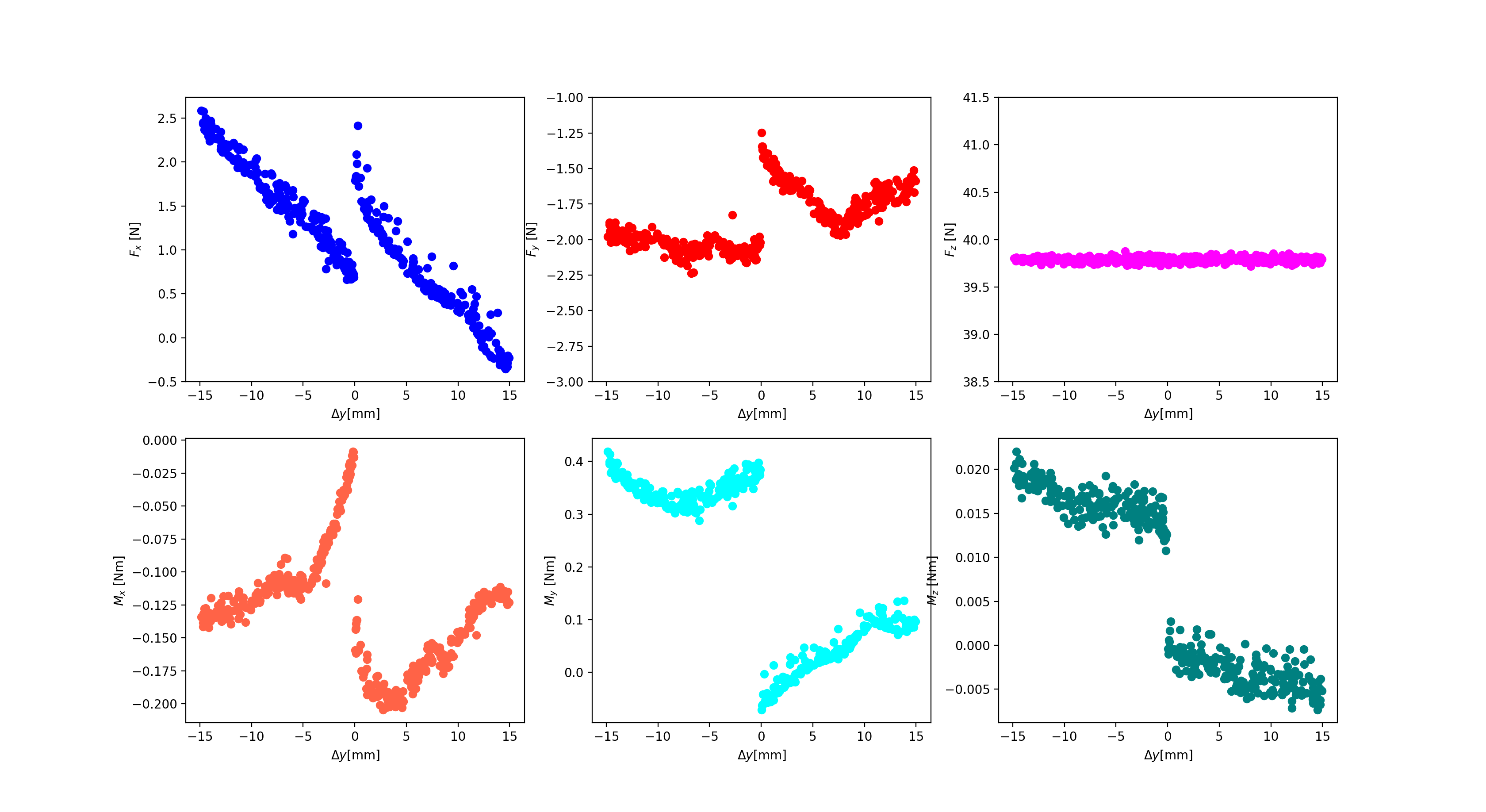} %
    \caption{Variation of contact wrench obtained from contact formations with different misalignment w.r.t. the hole along the $y$ axis, for a constant value of $x$. This plot shows that, in practice, we obtain a non-linear dependence between the misalignment and the wrench experienced during the contact formation.}
    \label{fig:force_w_error}
\end{figure*}

In this section, we present results of learning the corrective action in different contact configurations. In particular, we try to answer the following questions:
\begin{enumerate}
    \item Can the proposed method resolve the direction of the error as well as its magnitude?
    \item What is the range of misalignment that can be corrected using such a learned model?
    \item How well does the proposed controller perform on insertion tasks with inaccurate knowledge of goal the state?
\end{enumerate}

In the rest of this section, we try to answer these questions.

\subsection{Experiment Details}
We use a Mitsubishi Electric Factory Automation (MELFA) RV-$5$AS-D
 Assista $6$-DoF  arm (see Figure~\ref{fig:exp_setup}) for the experiments. The robot has a pose repeatability of $\pm 0.03$mm. The robot is equipped with Mitsubishi Electric F/T sensor $1$F-FS$001$-W$200$ (see Figure~\ref{fig:exp_setup}). The peg has a diameter of $20$ mm and the hole has a diameter of $22$ mm. We use the underlying stiffness controller of the robot to design the corresponding force controller. We use the stiffness controller with $K_z=2$ N/mm and $K_x=K_y=10$ N/mm as the stiffness constant during all the experiments. We use the GAC only in the axis of approach during insertion operation but one can implement it for all the six axes. 
 
 During the data collection procedure, error is introduced in the $x$ and $y$ position of the robot end-effector from the correct target state. This error is sampled from a uniform distribution in the range $[-R,R]$ around the circumference of the hole, where $R$ is the radius of the peg. Note that this is sufficiently larger than the error introduced in previous work, e.g.,~\cite{dong2021icra, Kulkarni2021LearningLearning}. For example, in~\cite{Kulkarni2021LearningLearning} the baseline success rate is already $40\%$ whereas in the present study, it is $0\%$. Furthermore, whereas the method in~\cite{Kulkarni2021LearningLearning} studied insertion in a slot-like environment (i.e., misalignment is only along one dimension), we study a more complex problem of peg insertion where error is introduced simultaneously in two dimensions. The robot is moved along the known approach direction with this error and using the force controller described in Section~\ref{sec:accommodation_controller_design} (see Figure~\ref{fig:proposed_method}). The controller maintains contact with the surface after the initial contact, and the force data for the entire interaction is stored. The total length of the trajectory is $30$ seconds, and the data is collected at $280$Hz, which is the default control rate of the robot.
 
 \subsection{Dependence of Contact Wrench on Misalignment}
In this section, we analyze the behavior of the contact wrench experienced by the peg during contact formation during an insertion attempt with varying amount of misalignment. The variation of the contact wrench with varying amount of misalignment along the $y$ axis, at a constant value of $x$, is shown in Figure~\ref{fig:force_w_error}. These plots show the (quasi-)steady state values of the contact wrench for a contact formation, that is, measured at the end of the insertion attempt. There are several things to infer from these plots shown in Figure~\ref{fig:force_w_error}. As seen in the plot of $F_z$, the force experienced in the vertical direction is constant for all values of misalignment. This force creates a moment about the x-axis which can be observed in the plot for $M_x$. This is also explained by a planar analytical model for the static equilibrium of the peg in the contact configuration. We also observe a moment about the y-axis of the peg which can be observed in the plot for $M_y$. We also observe that there is a non-linear dependence between the moments experienced by the peg and the amount of misalignment. While a planar analytical model would predict that the force and moments change linearly during these sustained contacts, we see here that there is significant non-linearity that is observed, and thus it provides the intuition for using a non-linear machine learning model for adjusting the misalignment based on the force readings obtained from the F/T sensor. 

\subsection{Predictive Model for Correcting Misalignment}
\begin{figure}
    \centering
    \includegraphics[width=0.40\textwidth]{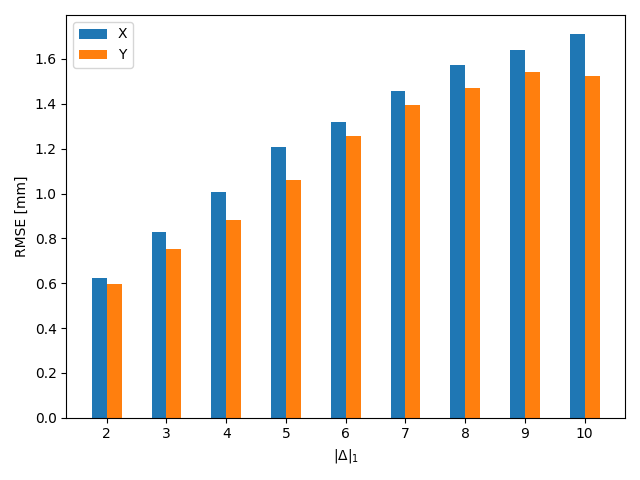} %
    \caption{Mean RMSE error (lower is better) for misalignment prediction along the $x$ and $y$ axis with increasing values of misalignment. As we can see from the plots, the RMSE monotonically increases with increasing misalignment.}
    \label{fig:rmse_plots}
\end{figure}

In this section, we describe learning a model to predict the actual misalignment between the hole and the peg. We train a models to predict the misalignment and try to understand the accuracy and degree to which the misalignment can be corrected using a supervised machine learning model. The experimental setup that we use for data collection is shown in Figure~\ref{fig:exp_setup} and was described in the previous section. We use Gaussian processes for learning the predictive model for misalignment. Gaussian processes are data efficient, and have been shown to perform well in a wide range of problems~\cite{9387127,romeres2019anomaly,romeres2019semiparametrical}. The Gaussian process regression is trained using the data collected using the accommodation controller described in Section~\ref{sec:accommodation_controller_design}. As was shown earlier in Figure~\ref{fig:accommodation_controller_plot}, the system reaches (quasi) steady state using the force controller. The Gaussian process regression is trained using the (quasi-)steady-state force signature obtained by the force controller. Thus the input to the regression model is a $6$ dimensional vector which is used to predict the misalignment along the $x$ and $y$ axis. We train two GPR models, one for each axis which are then used to predict the errors in each of the axis. We collect data from a total of $1,200$ trajectories with randomly sampled misalignment around the circumference of the hole. We train two different regression models to predict error along the x-axis and the y-axis. We use a sum-kernel with the RBF kernel and the white kernel to prevent overfitting of data. Figure~\ref{fig:rmse_plots} shows the prediction of the regression models for misalignment along the two axis. These plots show the mean RMSE predicted from $50$ different GP models trained and tested over different subsets of data from the total collected data. We also observe that the models have consistently better predictive behavior in the $y$ axis compared to the $x$ axis (this is an artifact of the contact configurations and not always true in general).  We also train Gaussian process classifiers to predict the direction of error using a similar sum-kernel to analyze the separation of data in predicting the direction. The classification results are listed in Table~\ref{table:GPC_prediction} which reflects almost perfect direction recovery. 

\begin{table}[t]
    \caption{{Classification accuracy in prediction of direction of misalignment along X and Y axis using Gaussian process classifiers.}}
    \centering
    \begin{tabular}{c|c}
     Axis & Classification Accuracy (higher is better)\\
         \hline\hline  X & 0.97\\
         \hline Y & 1.0 \\
    \end{tabular}
    \label{table:GPC_prediction}
\end{table}

\begin{figure}
    \centering
    \includegraphics[width=0.32\textwidth]{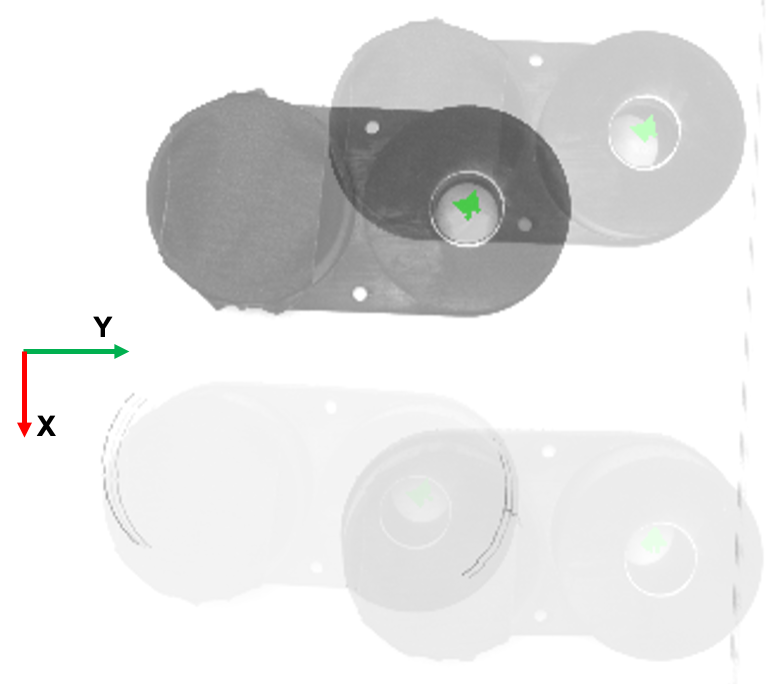} %
    \caption{The hole surface is moved in the plane in an approximate rectangular region of $12$ cm $\times$ $10$ cm. This figure shows the approximate region in the plane where the hole could be detected by the external camera (see Figure~\ref{fig:exp_setup}). Note that the images show the standard output of an industrial-grade camera. The frame shown is the robot frame of reference (see also Figure~\ref{fig:exp_setup}).}
    \label{fig:hole_detection}
\end{figure}

\subsection{Performance of the ML-based Controller}
In this section, we evaluate performance of the learning-based controller in correcting the misalignment during insertion experiments. We use the Gaussian process classification model described in the previous section to correct for the positional error using force feedback. In particular, we design a controller that uses a reference trajectory and the learned correction model to successfully perform insertion using force feedback. As shown in Figure~\ref{fig:proposed_method}, we use the accommodation controller at the lower level, and then use the predictions of the GP model to correct the reference trajectory using the prediction. We perform two sets of experiments. In the first set of experiments, the hole base is manually moved around in the plane (see Figure~\ref{fig:hole_detection}) inside a rectangular region of approximately $12$ cm $\times$ $10$ cm to generate $50$ novel goal locations. The hole location is detected by the external camera. This experiment tries to test the generalization of the controller to novel contact formations as the hole base is moved around. We observe that the camera error is biased in a particular direction, and thus the robot always ends up approaching the correction from a particular direction. Thus, we also design a second set of experiments where artificial errors are added to known location by sampling from the uniform distribution. This distribution is the same as during training time. During the force feedback correction, we use the predictions made by the Gaussian process classifier to move the robot in the direction predicted by the learned model. A unit step of $2$mm is used to move the robot along each axis. It can be adaptively changed based on the successive steps predicted by the model. The robot uses the contact wrench at the end of the contact trajectory as input to the learned Gaussian process model, uses the predictions from the model to make corrections in the $x$ and $y$ axis as predicted by the classifier model, makes contact with the hole surface again, and repeats the movement till either it achieves successful insertion or declares failure. A failure is declared based on either the total number of attempts exceeds the maximum number of attempts ($10$) or the controller diverging from the desired goal state more than $15$ mm. We implement two different controllers during test -- the first being the DMP trajectory (LfD) computed using the initial state and the novel goal location. Note that this is computed using a single successful demonstration directly provided on the robot using a joystick. The second controller is the proposed DMP trajectory augmented with the additional force-feedback controller (LfD+SL). The test results are listed in Table~\ref{table:controller_success_rate}. As can be seen, the proposed method achieves very high success rate during test times, while the DMP-based controller fails in all test condition in the final insertion.
\begin{table}[t]
    \caption{{Success rate of different controllers}}
    \centering
    \begin{tabular}{c|c|c}
     Experiment type & LfD+ SL & LfD \\
         \hline\hline  Moving the hole & 48/50 &0/50\\
         \hline Error added at known location & 79/80 & 0/80 \\
    \end{tabular}
    \label{table:controller_success_rate}
\end{table}


\section{Conclusions and Future Work}
Insertion is the most important operation of robotic assembly systems. Consequently, it has been widely studied to create robotic systems which can perform autonomous insertion using sensor-based feedback. However, the problem still remains open in several aspects, and there is no single technique to perform generalized insertion. In this paper, we presented a learning method for solving the problem of peg insertion during robotic assembly. Most of robotic assembly problems involve sustained contact phenomena, which makes it harder to ensure reliable and safe operation. We presented the design and analysis of an adaptive force controller that allows us to maintain sustained contacts with limited contact force in the presence of dynamic reference trajectory. We show that our proposed controller results in (quasi-)steady-state behavior by modifying the reference trajectory as the robot comes in contact with the external environment. We use this controller for safe exploration to collect data during the insertion process. This is used to learn a predictive model that can correct the misalignment during insertion. In our experiments, we show that we achieve approximately $98.75\%$ success rate with error sampled in the range of $[-R,R]$ in both the $x$ and $y$ axis, where $R$ is the radius of the inserted part. In contrast to the use of closed-form expressions used for solving similar problems in \cite{Gottschlich1989AMating}, our method does not need careful calibration, and works even when the axis of the peg is not aligned with the F/T sensor. Similarly, in comparison to the method recently reported in \cite{Kulkarni2021LearningLearning}, our method uses supervised learning instead of reinforcement learning, thus avoiding the need for prolonged random exploration.

Our method uses the quasi steady-state force data for learning the corrective compliance. However, converging to this quasi-steady state does take some time, which probably makes the insertion operation slower than it could be. In the future, we would like to use contact force readings from times before full convergence, hopefully speeding up the insertion process. In addition, we plan to study and analyze contact configurations with directional (rotational) misalignment between the peg and the hole.

\bibliographystyle{IEEEtran}
\bibliography{references}

\end{document}